\documentclass[pdflatex,sn-basic]{sn-jnl}

\jyear{2022}%

\usepackage{graphicx}
\usepackage{amsmath}
\usepackage{amssymb}
\usepackage{booktabs}
\usepackage{colortbl}
\usepackage{subcaption}

\usepackage[capitalize]{cleveref}
\crefname{section}{Sec.}{Secs.}
\Crefname{section}{Section}{Sections}
\Crefname{table}{Table}{Tables}
\crefname{table}{Tab.}{Tabs.}

\def\x{{\boldsymbol x}}
\def\y{{\boldsymbol y}}
\def\z{{\boldsymbol z}}
\def\u{{\boldsymbol u}}

\def\algo{{\textsc{TRAS}}}

\begin{document}

\title[Transfer and Share]{Transfer and Share: Semi-Supervised Learning from Long-Tailed Data}

\author*[1]{\fnm{Tong} \sur{Wei}}\email{weit@seu.edu.cn}
\equalcont{These authors contributed equally to this work.}

\author[2]{\fnm{Qian-Yu} \sur{Liu}}\email{\{liuqy,shijx\}@lamda.nju.edu.cn}
\equalcont{These authors contributed equally to this work.}

\author[2]{\fnm{Jiang-Xin} \sur{Shi}}

\author[3]{\fnm{Wei-Wei} \sur{Tu}}\email{tuweiwei@4paradigm.com}

\author*[2]{\fnm{Lan-Zhe} \sur{Guo}}\email{guolz@lamda.nju.edu.cn}

\affil[1]{\orgdiv{School of Computer Science and Engineering}, \orgname{Southeast University}, \orgaddress{\city{Nanjing}, \postcode{210096}, \country{China}}}

\affil[2]{\orgdiv{National Key Laboratory for Novel Software Technology}, \orgname{Nanjing University}, \orgaddress{ \city{Nanjing}, \postcode{210023}, \country{China}}}

\affil[3]{\orgname{4Paradigm Inc.}, \orgaddress{\city{Beijing}, \postcode{100000}, \country{China}}}


\abstract{Long-Tailed Semi-Supervised Learning (LTSSL) aims to learn from class-imbalanced data where only a few samples are annotated. Existing solutions typically require substantial cost to solve complex optimization problems, or class-balanced undersampling which can result in information loss. In this paper, we present the \textbf{\algo} (TRAnsfer and Share) to effectively utilize long-tailed semi-supervised data. \algo\ transforms the imbalanced pseudo-label distribution of a traditional SSL model via a delicate function to enhance the supervisory signals for minority classes. It then transfers the distribution to a target model such that the minority class will receive significant attention. Interestingly, \algo\ shows that more balanced pseudo-label distribution can substantially benefit minority-class training, instead of seeking to generate accurate pseudo-labels as in previous works. To simplify the approach, \algo\ merges the training of the traditional SSL model and the target model into a single procedure by sharing the feature extractor, where both classifiers help improve the representation learning. According to extensive experiments, \algo\ delivers much higher accuracy than state-of-the-art methods in the entire set of classes as well as minority classes.
}




\maketitle

\section{Introduction}\label{sec1}

\begin{figure}[h]%
\begin{subfigure}[b]{0.5\textwidth}
    \includegraphics[width=\textwidth]{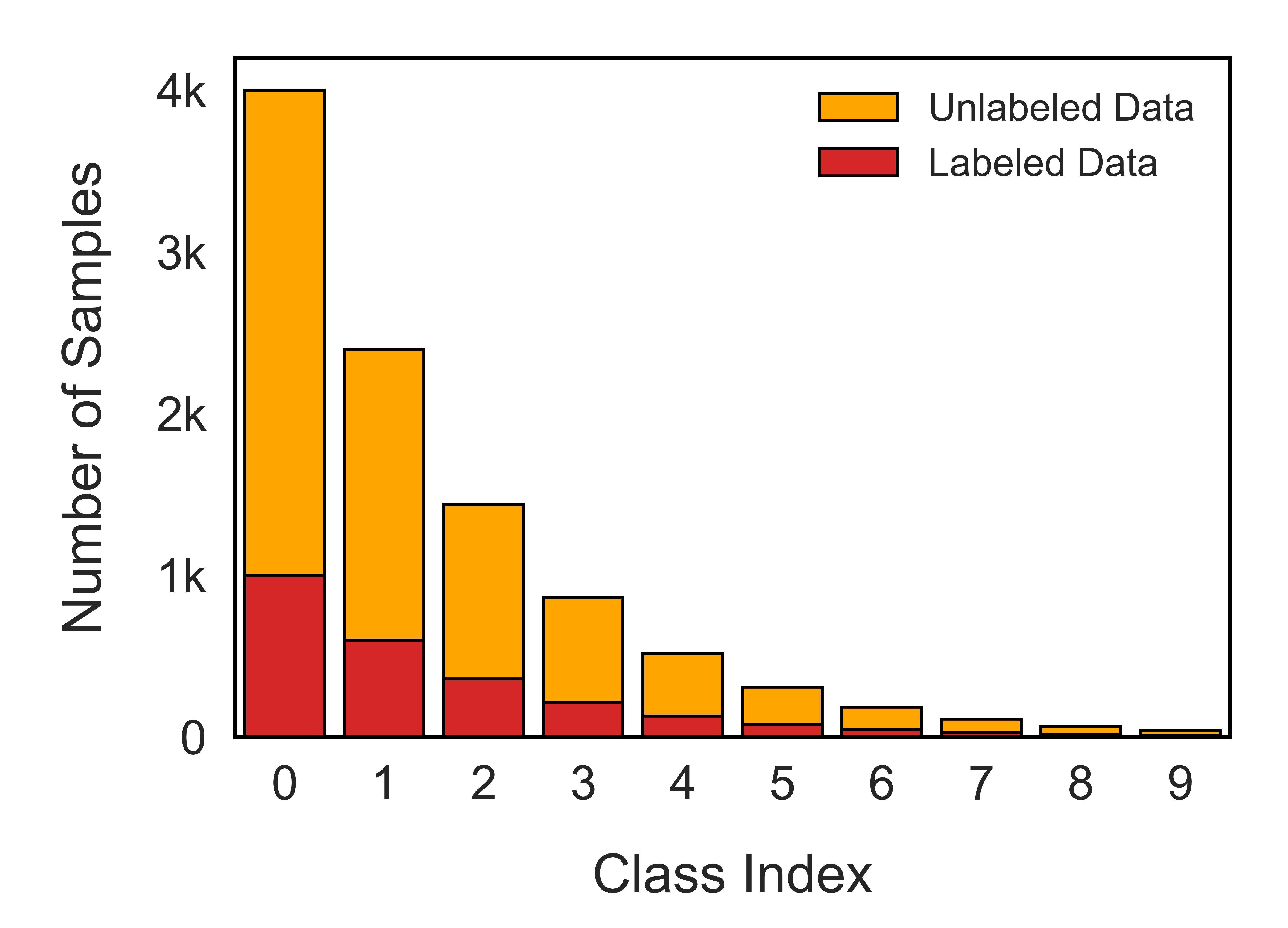}
    \caption{Long-tailed distribution}
    \label{a}
\end{subfigure}
\hfill
\begin{subfigure}[b]{0.48\textwidth}
    \includegraphics[width=\textwidth]{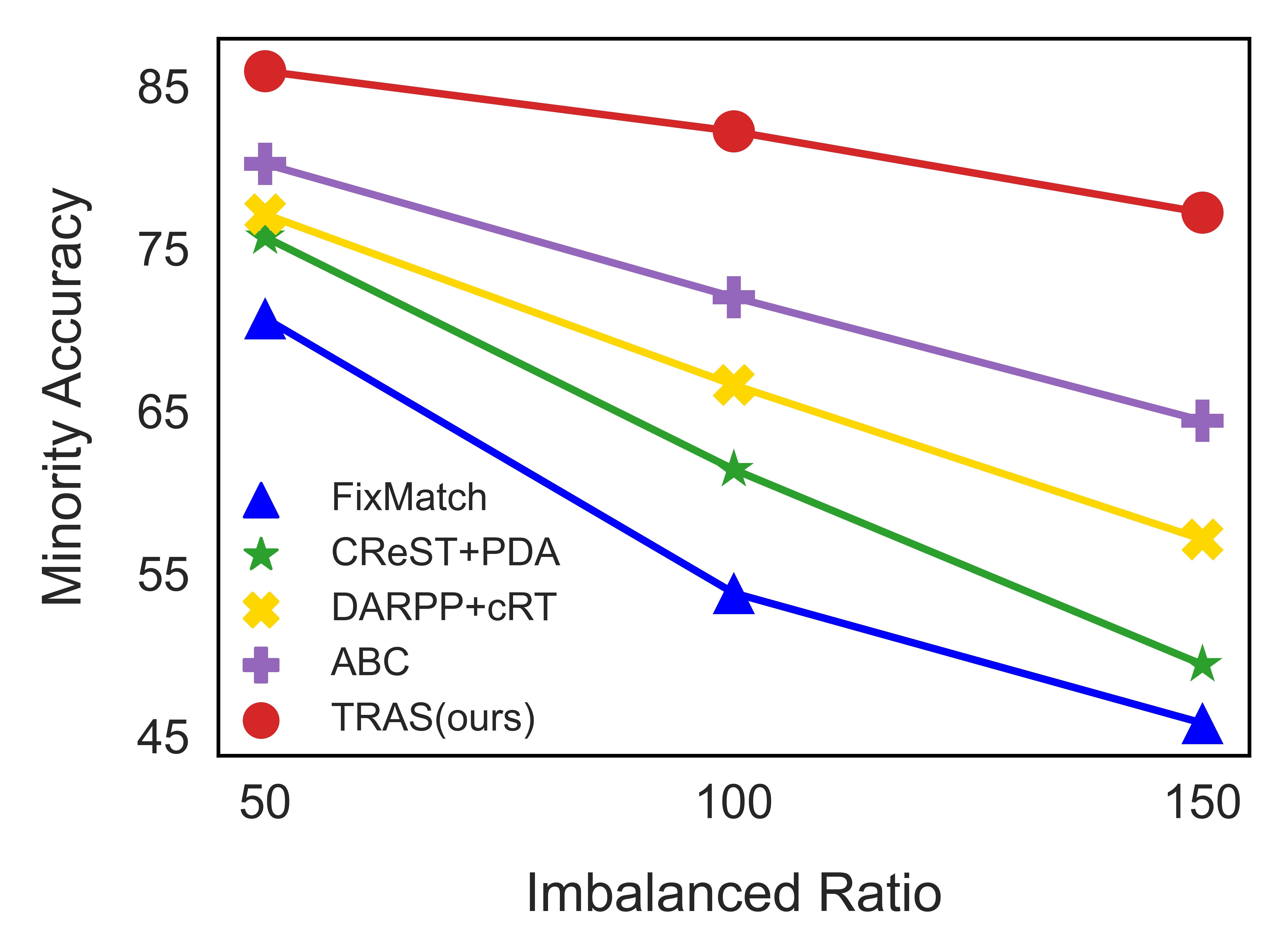}
    \caption{Minority-class accuracy}
    \label{b}
\end{subfigure}
\caption{(a) Long-tailed distribution of the training set under the main setting of CIFAR-10-LT. (b) Performance of minority-class accuracy(\%) on CIFAR-10-LT dataset under class imbalance ratio 50, 100, and 150 with 20\% of labels available. The proposed \algo\ heavily improves the minority-class accuracy.}\label{pro and acc}
\end{figure}

Deep Neural Networks (DNNs) have been successfully used in many real-world applications \citep{amodei2016deep,he2016deep}. However, the training of DNNs relies on \textit{large-scale} \& \textit{high-quality} datasets, which has become the core problem in practice. First, large-scale data annotation is highly expensive and only a small amount of labels can be accessed \citep{DBLP:conf/aaai/LiWWT19,DBLP:journals/ml/EngelenH20}. Second, collected data usually follows long-tailed distribution \citep{he2009learning,liu2019large,weit2020tnnls,Wei_2021_RoLT,DBLP:conf/pakdd/WeiSLZ22}, where only some classes (the majority class) have sufficient training samples while other classes (the minority class) own a few samples as shown in \Cref{a}.

To utilize unlabeled data, Semi-Supervised Learning (SSL) emerges as an interesting solution \citep{miyato2018virtual,tarvainen2017mean,berthelot2019mixmatch,sohn2020fixmatch,berthelot2019remixmatch,zhou2021step,guo2020safe}. It carries out model assumptions on the data distribution to build a learner to utilize unlabeled samples through selecting confident pseudo-labels. However, it is demonstrated that existing SSL methods tend to produce biased pseudo-labels towards the majority class \citep{kim2020distribution}, leading to undesirable performance.

Recently, Long-Tailed Semi-Supervised Learning (LTSSL) is proposed to improve the performance of SSL models on long-tailed data. The main ideas of existing LTSSL methods \citep{kim2020distribution,wei2021crest,lee2021abc} are two-fold. One is to improve the quality of pseudo-labels from the perspective of SSL. The other one is to employ class-balanced sampling or post-hoc classifier adjustment to alleviate class imbalance from the long tail perspective. These methods can improve the performance of conventional SSL models. However, the improvements are achieved with the cost of high computational overhead or losing information due to the undersampling of data.

How all data can be efficiently and effectively utilized is the core challenge of LTSSL and the focus of this paper. To this end, we propose a new method called \textbf{\algo} (TRAnsfer and Share) which has two key ingredients. \Cref{b} showcases the effectiveness of \algo\ in the minority class.

First, we compensate for the minority-class training by generating a more balanced pseudo-label distribution. Under the guidance of pseudo-label distribution, DNNs can mine the interaction information between classes to obtain richer information for minority classes. The idea of learning from label distribution has been explored in previous literature, such as label distribution learning \citep{geng2016label,gao2017deep,wang2019classification} and knowledge distillation \citep{xiang2020learning,he2021distilling,iscen2021cbd}, which however is still underexplored in LTSSL. To generate label distribution, knowledge distillation is a common approach via well-trained teacher models. 
Such a teacher model is not always available for SSL models because of limited long-tailed labeled data and high computation overhead.
Alternatively, we employ a \textit{conventional} SSL model with normally low accuracy on the minority class. This conventional SSL model is able to teach the learning of the student model after applying our proposed logit transformation. 
This transformation is particularly designed to enhance the minority-class \textit{supervisory signals} without introducing extra computational cost. 
Subsequently, through training the student model by imitating the enhanced supervisory signals, the minority class will receive significant attention.

Second, we propose to merge the training of teacher and student models as a single procedure to reduce the computational cost. To this end, we use a double-branch neural network with a shared feature extractor and two classifiers for producing the predictions of the teacher and student. The neural network is then trained in an end-to-end way by a joint objective of these two classifiers. In addition to reduce training cost and simplify the approach, we empirically find that both classifiers can help improve the representation learning and learn clear classification boundaries between classes. 

Our main contributions are summarized as follows:
\begin{enumerate}
    \item A new LTSSL method \algo\ is proposed, which significantly improves the minority-class training without introducing extra training cost.
    \item \algo\ transfers pseudo-label distribution from a vanilla SSL network (teacher) to another network (student) via a new logit transformation, instead of trying hard to construct a sophisticated LTSSL teacher model.
    \item \algo\ reveals the importance of the balancedness of pseudo-label distribution in transfer for LTSSL. 
    \item \algo\ merges the training of teacher and student models by sharing the feature extractor, which simplifies the training procedure and benefits the representation learning.
    \item  \algo\ achieves state-of-the-art performance in various experiments. Particularly, it improves minority-class performance by about 7\% in accuracy.
\end{enumerate}

\section{Related Work}\label{sec2}
\textbf{Semi-supervised learning.} Existing SSL methods aim to use unlabeled data to improve the generalization. For this purpose, consistency regularization and entropy minimization have become the most frequently used techniques and demonstrate considerable performance improvements.
Specifically, Mean-Teacher \citep{tarvainen2017mean} imposes consistency regularization between the prediction of the current model and the self-ensembled model obtained using exponential moving average. Virtual Adversarial Training (VAT) \citep{miyato2018virtual} encourages the model to minimize the discrepancy of model predictions for unlabeled data before and after applying adversarial perturbation. MixMatch \citep{berthelot2019mixmatch} minimizes the entropy of model predictions by sharpening the pseudo-label distribution. ReMixMatch \citep{berthelot2019remixmatch} improves MixMatch by imposing another distribution alignment regularizer and augmentation anchoring.
FixMatch \citep{sohn2020fixmatch} merges consistency regularization and entropy minimization by regularizing the prediction for weakly augmented and strongly augmented unlabeled data. However, the above-mentioned methods assume both labeled and unlabeled data is both class-balanced, leading to poor performance on the minority class when working on long-tailed datasets.

\textbf{Long-tailed semi-supervised learning.} To deal with long-tailed datasets, several LTSSL methods have been proposed. In a nutshell, exiting methods aim to select not only confident but also more class-balanced pseudo-labels to improve the generalization for minority classes. For instance, DARP \citep{kim2020distribution} proposes to estimate the underlying class distribution of unlabeled data, which is used to regularize the distribution of pseudo-labels. To this end, a convex optimization problem is solved. Additionally, CReST \citep{wei2021crest} 
proposes to use class-aware confidence thresholds for selecting more pseudo-labels for the minority class. Recently, ABC \citep{lee2021abc} proposes to use an auxiliary balanced classifier built upon a conventional SSL model by class-balanced undersampling. However, these approaches either suffer from high computational cost or loss of supervisory information. In this work, we propose a new algorithm \algo, which can fully utilize not only supervised data but also unsupervised data through efficient pseudo-label distribution transfer, and greatly improves the performance of the minority class.

\section{Method:~\algo}\label{sec3}

\begin{figure}[!t]%
    \includegraphics[width=0.9\textwidth]{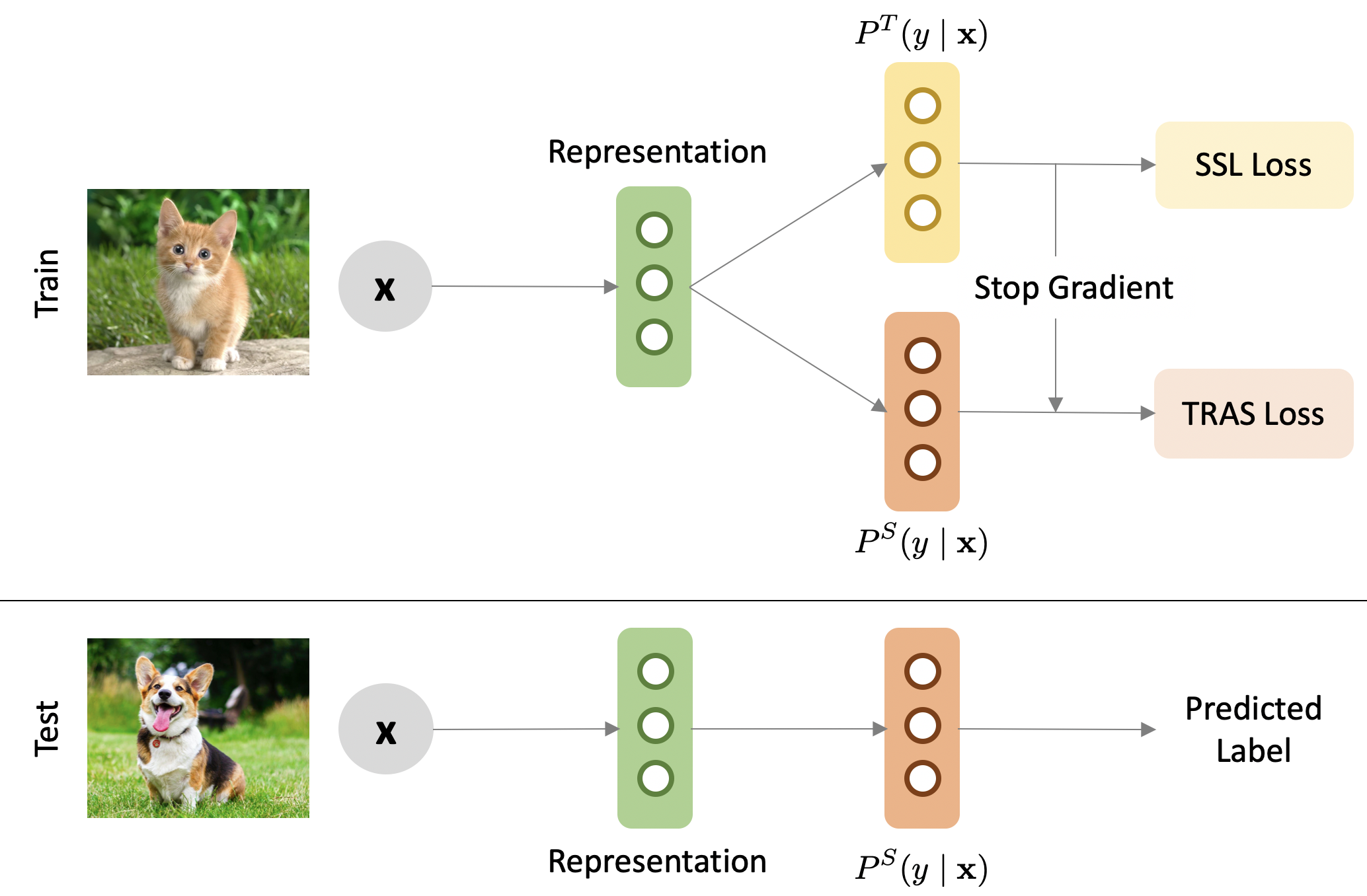}
    \caption{
		The \algo\ method in diagrammatic form. 
    } \label{fig:framework}
\end{figure}

We now introduce the problem setting in \Cref{problem-setting} and develop our proposed method \algo, which consists of two key ingredients described in \Cref{sec:ing1} and \Cref{sec:ing2}. \Cref{fig:framework} shows the framework of the proposed \algo.

\subsection{Problem Setting}\label{problem-setting}
Let $\mathcal{X}=\{(\x_i,y_i)\}_{i=1}^N$ be a labeled dataset, where $\x_i \in \mathbb{R}^d$ denotes the training example and $y_i \in \mathbb{R}$ is the corresponding label. We introduce an unlabeled dataset $\mathcal{U} = \{(\u_i)\}_{i=1}^M$ where $\u_i \in \mathbb{R}^d$ is the unlabeled data point. 
Following ABC \citep{lee2021abc}, we assume that the class distributions of $\mathcal{X}$ and $\mathcal{U}$ are identical. 
We denote the number of labeled data points of class $l$ as $N_l$ (notice that $\textstyle\sum_{l=1}^L N_l=N$), assuming that all classes are sorted by cardinality in descending order $N_1 \textgreater N_2 \textgreater \cdots \textgreater N_L$. Corresponding to LTSSL, we set the ratio of labeled data as $\beta =\frac{N}{N+M}$ and the ratio of the class imbalance as $\gamma=\frac{N_1}{N_L}$. 
Following previous LTSSL works, we divide the class space into the majority class and the minority class according to their frequencies in the training data.
Finally, our goal is to learn a model which generalizes well on both the majority class and the minority class.

Our proposed method, \algo, consists of a shared feature extractor and two classifiers, providing predictions for the teacher model $P^T(y\mid \x)$ and student model $P^S(y\mid \x)$. 
There are two key ingredients to \algo: (1) Learn through imitation, in which the student model imitates the adjusted output of the teacher model, and (2) transfer via sharing weights. In the following, we present technical details of these two ingredients.

\subsection{Ingredient \#1: Learn through Imitation}\label{sec:ing1}

Given labeled data, a typical approach is to train a classifier $f$ by optimizing the softmax cross-entropy:
\begin{equation}
\ell_{\text{CE}}(y, f(\x))=-\log \frac{e^{f_{y}(\x)}}{\sum_{y^{\prime} \in[L]} e^{f_{y^{\prime}}(\x)}}.
\end{equation}

In LTSSL, however, the distribution of labeled data is heavily class-imbalanced, such that the learned classifier would be biased towards the majority class. To improve the training of the minority class, we propose to use the distribution-aware cross-entropy loss:
\begin{equation}
\ell_{\text{DA-CE}}(y, f(\x))=-\log \frac{e^{f_{y}(\x)+\tau \cdot \log \pi_{y}}}{\sum_{y^{\prime} \in[L]} e^{f_{y^{\prime}}(\x)+\tau \cdot \log \pi_{y^{\prime}}}},
\end{equation}
where $\pi_y$ is the estimate of class prior $\mathbb{P}(y)$ and $\tau>0$ is a scaling parameter. By minimizing $\ell_{\text{DA-CE}}$, it encourages large margins between the true label and other negative labels. Using distribution-aware cross-entropy is not a new idea in the literature of long-tailed learning, such as Logit Adjustment \citep{menon2020long} and Balanced Softmax \citep{ren2020balanced}. Interestingly, existing methods show that the scaling parameter $\tau$ plays an important role in model training, but it is usually used as a constant, e.g., $\tau=1$. In the following, we show a new instance-dependent logit scaling method. 

In addition to a handful of labeled data, we can access to a large amount of unlabeled data to help improve the generalization. In LTSSL, the underlying distribution of unlabeled data is also long-tailed, and conventional SSL methods have shown impaired performance on the minority class.
This paper proposes to train the model using pseudo-label distribution, rather than biased one-hot pseudo-labels. Intuitively, label distribution offers more supervisory signals and can benefit the minority-class training.
We generate pseudo-label distribution by first training a vanilla SSL model as the teacher, and then training a student model by imitating the output distribution of teacher model. We opt for minimizing their Kullback-Leibler (KL) divergence:
\begin{equation}
    \ell_{\text{KL}}\left( \tilde{\y}^{T} , \tilde{\y}^{S}\right) = \sum_{l=1}^{L} \tilde{y}^{T}_{l} \log \frac{\tilde{y}^{T}_{l}}{\tilde{y}^{S}_{l}},
\end{equation}
where $\tilde{\y}^T$ and $\tilde{\y}^S$ are output probabilities of the teacher and student model respectively, which illustrate the implicit information of label distribution.

\begin{figure}[h]%
  \begin{minipage}[c]{0.55\textwidth}
    \includegraphics[width=\textwidth]{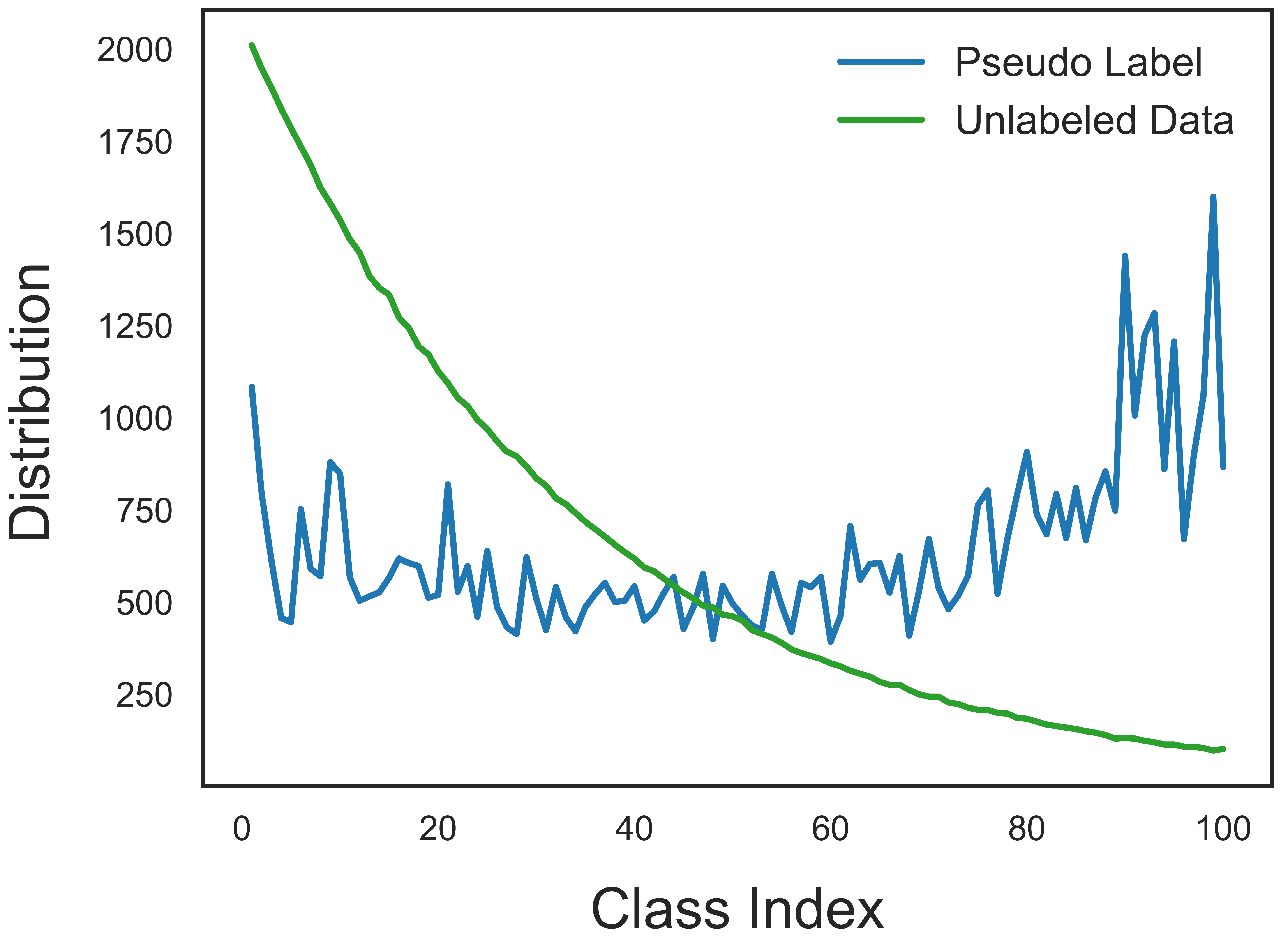}
  \end{minipage}\hfill
  \begin{minipage}[c]{0.43\textwidth}
    \caption{
		Comparison of ground-truth label distribution and our generated pseudo-label distribution on CIFAR-100-LT dataset under class imbalance ratio 20 with 40\% of labels available.
    } \label{fig:pseudo-label-distribution}
  \end{minipage}
\end{figure}

Note that the teacher model is trained via a conventional SSL algorithm and the produced pseudo-label distribution is still biased towards the majority class. To further enhance supervisory signals for the minority class, we present a new logit transformation to adjust the output of the teacher model. Specifically, for sample $\x$, we transform its pseudo-label distribution as follows:
\begin{equation}
    \tilde{\y}^T = \text{softmax}\left(\phi \left(\z^T\right) \right) = \text{softmax} \left( \z^T -\tau\left(\hat{y}\right) \cdot \log \boldsymbol{\pi} \right),
\end{equation}
where $\z^T$ is the output logits and $\hat{y}$ is the pseudo-label of $\x$. In this way, the pseudo-label distribution of unlabeled data is more balanced than the original distribution. We demonstrate the generated label distribution in \Cref{fig:pseudo-label-distribution}. 

Notably, different from previous works that treat $\tau$ as a constant to scale the output logits, we use $\tau$ as a function of pseudo-labels. Concretely, given the pseudo-label $\hat{y}$, we define $\tau(\hat{y}) =A \cdot \alpha_{\hat{y}} + B$, where $\boldsymbol{\alpha} = \text{softmax} (-\log \boldsymbol{\pi})$ is a $\hat{y}$-dependent function, $A$ and $B$ are constants. This is because adjusting pseudo-label distribution to over-compensate the minority class can be harmful to the majority class. By employing the $\hat{y}$-dependent logit transformation function, we can alleviate this problem by flattening the label distribution of predicted minority-class samples more aggressively than other samples.
In experiments, we simply set $A=B=2$. Applying the proposed logit transformation generates a more balanced pseudo-label distribution to improve the training of the minority class as in \Cref{fig:pseudo-label-distribution}. 

Putting together the objectives for labeled and unlabeled data, we minimize the loss function for \algo\ as follows:
\begin{equation}
    \mathcal{L}_{\text{\algo}} =  \underbrace{\sum_{i=1}^{N} \ell_{\text{DA-CE}}\left(y_i, \z^S_i\right)}_{\text{supervised\ loss}} + \underbrace{\sum_{j=1}^{M} \mathbb{I}\left(\max\left(\tilde{\y}^S_j\right) \geq t\right) \ell_{\text{KL}}\left( \tilde{\y}^T_j, \tilde{\y}^S_j\right)}_{\text{unsupervised\ loss}}.
\end{equation}
Here, $\mathbb{I}(\cdot)$ is the indicator function, $t$ denotes the confidence threshold and we adopt the common setup $t = 0.95$ for confident pseudo-labels from the student.

In this way, pseudo-label distribution can naturally describe the implicit information between labels. By applying the logit transformation, the distribution encodes more informative supervisory signals for the minority class. By imitating pseudo-label distribution, the student can alleviate data scarcity for minority classes.

\subsection{Ingredient \#2: Transfer via Sharing Weights}\label{sec:ing2}

Learning through imitation of the teacher model can significantly compensate for the training of the minority class, however, it needs to train two DNNs models sequentially and is costly in SSL. 

To reduce the time consumption and simplify the approach, we propose to merge the training of teacher and student models into a single training procedure. In other words, the teacher and student share the feature extractor network. 
We further partition the parameter space into three disjoint subsets; (1) Let $\psi(\x)$ be a feature extractor for $\x$. (2) Let $f^T(\psi(\x))$ be a teacher classifier producing the prediction $\tilde{\y}^T$. (3) Similarly, let $f^S(\psi(\x))$ be a student classifier producing the prediction $\tilde{\y}^S$. Subsequently, let us define:
\begin{equation}
    \z^T = \text{stop\_gradient}\left(f^T\left(\psi(\x)\right)\right),
\end{equation}
which is the output logits of the teacher model except that its gradient will not be calculated to update the teacher model's classifier weights. Recall that function $\phi(\cdot)$ acts as a logit transformer of $\z^T$, we then consider:
\begin{equation}
    \mathcal{L}_{\text{\algo}} = \underbrace{\sum_{i=1}^{N} \ell_{\text{DA-CE}}\left(y_i, \z^S_i\right)}_{\text{supervised\ loss}} + \underbrace{\sum_{j=1}^{M} \mathbb{I}\left(\max\left(\tilde{\y}^S_j\right) \geq t\right) \ell_{\text{KL}}\left( \text{softmax} \left( \phi \left(\z^T_j\right)\right), \tilde{\y}^S_j\right)}_{\text{unsupervised\ loss}},
\end{equation}
as the joint objective.
Note that the teacher and student share a single feature extractor, it only adds a linear classifier to the conventional SSL model, which incurs negligible training cost.

Let $\mathcal{L}_{\text{SSL}}$ denote the loss for a conventional SSL method, the total loss function that \algo\ optimizes is:
\begin{equation}
     \mathcal{L}_{\text{Total}} = \mathcal{L}_{\text{\algo}} + \mathcal{L}_{\text{SSL}}.
\end{equation}
Particularly, if FixMatch is employed as the teacher model, $\mathcal{L}_{\text{SSL}}$ consists of a cross-entropy loss on labeled data and a consistency regularization on unlabeled data. Specifically, we have:
\begin{equation}
    \mathcal{L}_{\text{SSL}} =
    \underbrace{\sum_{i=1}^{N} \ell_{\text{CE}}\left(y_i, \z^T_i\right)}_{\text{supervised\ loss}} + \underbrace{\sum_{j=1}^{M} \mathbb{I}\left(\max\left(\z^T_j\right) \geq t\right) \ell_{\text{CE}}\left(\hat{y}_j,
    \tilde{\z}^T_j\right)}_{\text{unsupervised\ loss}},
\end{equation}
where $\z^T$ and $\tilde{\z}^T$ are the output logits for weak and strong data augmentation, $\hat{y}=\arg \max_l z^T_l$ represents the pseudo-label for unlabeled data.
In inference, we use the student classifier $f^S$ to predict the label.

\subsection{Connection to Previous Work}

One may note that the basic idea of our \algo\ can transfer knowledge distribution from a vanilla teacher model to a student model that has good generalization for the minority class. The technique is related to knowledge distillation which has been explored in some recent long-tailed learning works. For instance, LFME \citep{xiang2020learning} proposes to train the student model via distilling multiple teachers trained on less imbalanced datasets. DiVE \citep{he2021distilling} shows that flattening the output distribution of the teacher model using a constant temperature parameter can help the learning of minority classes. CBD \citep{iscen2021cbd} distills features from the teacher to the student and shows that it can improve the learned representation of the minority class. Last but not least, xERM \citep{beierxERM} obtains an unbiased model by properly adjusting the weights between empirical loss and knowledge distillation loss.

In contrast to previous works that aim to solve supervised long-tailed learning, this paper studies semi-supervised long-tailed learning, where the amount of labeled data is much more limited. Moreover, previous works need to train teacher models via well-established long-tailed learning methods. However, our method \algo\ only needs a vanilla SSL model as a teacher. Additionally,  these methods have multiple-stage training procedures, but our method is simpler and can be trained in an end-to-end way.

\section{Experiments}\label{sec4}

We conduct experiments on long-tailed version of CIFAR-10, CIFAR-100, and SVHN, in comparison with state-of-the-art LTSSL methods. We then perform hyper-parameter sensitivity studies and ablation studies to better understand our proposed \algo.

\subsection{Experimental Setup}\label{subsec1}

\textbf{Datasets.} We conduct experiments on common datasets long-tailed CIFAR-10 (CIFAR-10-LT), long-tailed CIFAR-100 (CIFAR-100-LT) and long-tailed SVHN (SVHN-LT) to evaluate our method. Without loss of generality, for imbalanced SSL settings, we randomly resample the datasets to meet the assumption that the distribution of labeled and unlabeled samples is consistent. We set the ratio of the class imbalance as $\gamma$  ($\gamma=\frac{N_1}{N_L}$) and the number of labeled data points of class $l$ as $N_l$, where $N_l= N_1 * \gamma^{-{{l-1}\over{L-1}}}$ and $M_l$ for the unlabeled. Specifically, we set $N_1+M_1=5000$, $L = 10$ for CIFAR-10-LT and SVHN-LT, $N_1+M_1=500$, $L = 100$ for CIFAR-100-LT respectively. 

Following the previous work  \citep{lee2021abc}, we evaluate the classification performance with imbalance ratio $\gamma$ = 100 and 150 for CIFAR-10-LT and SVHN-LT and $\gamma$ = 20 and 30 for CIFAR-100-LT. The ratio of labeled data $\beta$ is 10\%, 20\% and 30\% for CIFAR-10-LT and SVHN-LT, 20\%, 40\% and 50\% for CIFAR-100-LT. Since the test set remains balanced, overall accuracy, minority-class accuracy, and Geometric Mean scores (GM) \citep{branco2016survey} with class-wise sensitivity are three main metrics to validate the proposed method.

\textbf{Setup}. We implement our method with FixMatch over the backbone of Wide ResNet-28-2 \citep{zagoruyko2016wide}. Our method is compared with the supervised baseline, long-tailed supervised learning methods, and long-tailed semi-supervised learning methods, denoted by (a) Vanilla; (b) VAT \citep{miyato2018virtual} and FixMatch \citep{sohn2020fixmatch}; (c) BALMS \citep{ren2020balanced}, classifier Re-Training (cRT) \citep{kang2019decoupling}; (d) DARP \citep{kim2020distribution}, CReST \citep{wei2021crest}, ABC \citep{lee2021abc}. We set the hyper-parameters by following FixMatch and train the neural networks for $500$ epochs with $500$ mini-batches in each epoch, with the batch size of $64$, using Adam optimizer \citep{kingma2015adam}. The learning rate is $0.002$ with a decay rate of $0.999$. We start optimizing \algo\ after training FixMatch for $10$ epochs. For all experiments, we report the mean and standard deviation of test accuracy over multiple runs.

\subsection{Experimental Results}

First, the performance of the algorithms compared under the main setting is in Table~\ref{tab1}. Results of related methods are borrowed from ABC \citep{lee2021abc}. It can been see that our method achieves the best performance, and the improvement on the minority class is impressive. It is known that normal SSL methods such as VAT and FixMatch perform unsatisfactorily on the minority class because pseudo-labels of unlabeled data are affected by the biased model thus hindering the learning of minority classes. 
Our method significantly improves the performance on the minority class by exploiting knowledge transfer to generate balanced label distribution, which conveys more implicit information than the one-hot pseudo-labels used in most previous LTSSL works. Moreover, our standard deviation is lower than other LTSSL methods, showing the superior stability of \algo.

\begin{table}[h]
\begin{center}
\caption{Overall accuracy(\%)/minority-class accuracy(\%) under the main setting.}\label{tab1}%
\begin{tabular}{lccc}
\toprule
  & CIFAR-10-LT  & SVHN-LT &  CIFAR-100-LT\\
\midrule
Algorithm & $\gamma = 100, \beta = 20\%$ & $\gamma = 100, \beta = 20\%$ & $\gamma = 20, \beta = 40\%$ \\
\midrule
Vanilla    & 55.3$_{\pm1.30}$/33.9$_{\pm1.88}$  & 77.0$_{\pm0.67}$/63.3$_{\pm1.25}$  & 40.1$_{\pm1.15}$/25.2$_{\pm0.95}$ \\
VAT    & 55.3$_{\pm0.88}$/28.2$_{\pm1.55}$   & 81.3$_{\pm0.47}$/68.2$_{\pm0.88}$  & 40.4$_{\pm0.34}$/24.8$_{\pm0.38}$  \\
BALMS    & 70.7$_{\pm0.59}$/69.8$_{\pm1.03}$  & 87.6$_{\pm0.53}$/85.0$_{\pm0.67}$  & 50.2$_{\pm0.54}$/42.9$_{\pm1.03}$\\
\midrule
FixMatch & 72.3$_{\pm0.33}$/53.8$_{\pm0.63}$ & 88.0$_{\pm0.30}$/79.4$_{\pm0.54}$ & 51.0$_{\pm0.20}$/32.8$_{\pm0.41}$  \\
w/ CReST+PDA & 76.6$_{\pm0.46}$/61.4$_{\pm0.85}$ & 89.1$_{\pm0.69}$/81.7$_{\pm1.18}$ & 51.6$_{\pm0.29}$/36.4$_{\pm0.46}$  \\
w/ DARP & 73.7$_{\pm0.98}$/57.0$_{\pm2.12}$ & 88.6$_{\pm0.19}$/80.5$_{\pm0.54}$ & 51.4$_{\pm0.37}$/33.9$_{\pm0.77}$  \\
w/ DARP+cRT  & 78.1$_{\pm0.89}$/66.6$_{\pm1.55}$ & 89.9$_{\pm0.44}$/83.5$_{\pm0.61}$ & 54.7$_{\pm0.46}$/41.2$_{\pm0.42}$  \\
w/ ABC  & 81.1$_{\pm0.82}$/72.0$_{\pm1.77}$ & 92.0$_{\pm0.38}$/87.9$_{\pm0.73}$ & 56.3$_{\pm0.19}$/43.4$_{\pm0.42}$ \\
\rowcolor[gray]{0.8} w/ \algo(ours)  & \textbf{84.3}$_{\pm0.25}$/\textbf{82.2}$_{\pm0.44}$  & \textbf{93.4}$_{\pm0.51}$/\textbf{92.5}$_{\pm0.26}$ & \textbf{58.5}$_{\pm0.17}$/\textbf{50.3}$_{\pm0.22}$ \\
\botrule
\end{tabular}
\end{center}
\end{table}

To further validate the effectiveness of our method, we report the performance on various settings. The results on CIFAR-10-LT, SVHN-LT and CIFAR-100-LT are reported in~\Cref{tab2}, \Cref{tab3} and \Cref{tab4}. Our method \algo\ outperforms other methods in all cases with respect to both overall accuracy and minority-class accuracy. Particularly, \algo\ achieves about 10\%, 5\%, 7\% improvements in the minority class on three datasets. Moreover, \algo\ is more robust to class imbalance. As the imbalance ratio increases, existing methods severely deteriorate their performance, while the accuracy of our method drops slightly.

\begin{table}[h]
\begin{center}
\caption{Overall accuracy(\%)/minority-class accuracy(\%) for CIFAR-10-LT. Two imbalance ratios $\gamma$ and three labeled data ratios $\beta$ are evaluated.}\label{tab2}%
\begin{tabular}{lcccccc}
\toprule
\multicolumn{4}{c}{CIFAR-10-LT}  \\
\midrule
Algorithm & $\gamma= 100, \beta= 10\%$  & $\gamma= 100, \beta= 30\%$ & $\gamma= 150, \beta= 20\%$ \\
\midrule
FixMatch & 70.0$_{\pm0.59}$/48.9$_{\pm1.04}$ & 74.9$_{\pm0.63}$/58.2$_{\pm1.28}$ &  68.5$_{\pm0.60}$/45.8$_{\pm1.15}$  \\
w/ CReST+PDA & 73.9$_{\pm0.40}$/58.9$_{\pm1.14}$ & 77.6$_{\pm0.73}$/64.0$_{\pm1.39}$ & 70.0$_{\pm0.82}$/49.4$_{\pm1.52}$  \\
w/ DARP+cRT  & 74.6$_{\pm0.98}$/59.2$_{\pm2.12}$ & 79.0$_{\pm0.25}$/67.7$_{\pm0.95}$ & 73.2$_{\pm0.85}$/57.1$_{\pm1.13}$  \\
w/ ABC  & 77.2$_{\pm1.60}$/65.7$_{\pm2.85}$ & 81.5$_{\pm0.29}$/72.9$_{\pm0.96}$ & 77.1$_{\pm0.46}$/64.4$_{\pm0.92}$ \\
\rowcolor[gray]{0.8} w/ \algo(ours)  & \textbf{82.1}$_{\pm0.41}$/\textbf{78.6}$_{\pm0.51}$  & \textbf{85.0}$_{\pm0.46}$/\textbf{83.0}$_{\pm0.85}$ & \textbf{81.7}$_{\pm0.44}$/\textbf{77.2}$_{\pm0.92}$ \\
\botrule
\end{tabular}
\end{center}
\end{table}

\begin{table}[!h]
\begin{center}
\caption{Overall accuracy(\%)/minority-class accuracy(\%) on SVHN-LT. Two imbalance ratios $\gamma$ and three labeled data ratios $\beta$ are evaluated.}\label{tab3}%
\begin{tabular}{lcccccc}
\toprule
\multicolumn{4}{c}{SVHN-LT}  \\
\midrule
Algorithm & $\gamma= 100, \beta= 10\%$  & $\gamma= 100, \beta= 30\%$  & $\gamma= 150, \beta= 20\%$ \\
\midrule
FixMatch & 88.5$_{\pm0.25}$/80.3$_{\pm0.42}$ & 88.7$_{\pm0.36}$/80.7$_{\pm0.65}$ & 85.6$_{\pm0.17}$/74.6$_{\pm0.43}$  \\
w/ CReST+PDA & 89.2$_{\pm0.43}$/81.7$_{\pm0.95}$ & 89.9$_{\pm0.36}$/83.0$_{\pm0.37}$ & 86.7$_{\pm0.89}$/76.7$_{\pm1.70}$  \\
w/ DARP+cRT  & 89.3$_{\pm0.33}$/83.9$_{\pm0.47}$ & 90.7$_{\pm0.28}$/84.8$_{\pm0.37}$ & 88.0$_{\pm0.74}$/80.1$_{\pm1.88}$  \\
w/ ABC  & 92.3$_{\pm0.38}$/88.7$_{\pm0.92}$ & 92.3$_{\pm0.34}$/88.3$_{\pm0.49}$ &  91.2$_{\pm0.15}$/86.2$_{\pm0.15}$ \\
\rowcolor[gray]{0.8} w/ \algo(ours)  & \textbf{93.2}$_{\pm0.22}$/\textbf{92.5}$_{\pm0.10}$  & \textbf{93.9}$_{\pm0.20}$/\textbf{93.4}$_{\pm0.33}$ &  \textbf{92.1}$_{\pm0.36}$/\textbf{91.1}$_{\pm0.86}$ \\
\botrule
\end{tabular}
\end{center}
\end{table}

\begin{table}[!h]
\begin{center}
\caption{Overall accuracy(\%)/minority-class accuracy(\%) on CIFAR-100-LT. Two imbalance ratios $\gamma$ and three labeled data ratios $\beta$ are evaluated.}\label{tab4}%
\begin{tabular}{lcccc}
\toprule
\multicolumn{4}{c}{CIFAR-100-LT}     \\
\midrule
Algorithm & $\gamma= 20, \beta= 20\%$  & $\gamma= 20, \beta= 50\%$ & $\gamma= 30, \beta= 40\%$ \\
\midrule
FixMatch & 46.1$_{\pm0.23}$/26.6$_{\pm0.34}$ & 52.3$_{\pm0.54}$/34.7$_{\pm0.80}$ & 47.6$_{\pm0.09}$/27.6$_{\pm0.21}$  \\
w/ CReST+PDA & 46.7$_{\pm0.49}$/29.3$_{\pm0.54}$ & 52.7$_{\pm0.06}$/37.4$_{\pm0.37}$ &48.5$_{\pm0.06}$/30.0$_{\pm0.04}$  \\
w/ DARP+cRT  & 48.9$_{\pm0.11}$/33.5$_{\pm0.17}$ & 55.9$_{\pm0.43}$/43.5$_{\pm1.28}$ & 51.3$_{\pm0.29}$/36.4$_{\pm0.50}$  \\
 w/ ABC  & 49.7$_{\pm0.40}$/34.6$_{\pm0.76}$ & 58.3$_{\pm0.74}$/46.7$_{\pm1.12}$& 53.6$_{\pm0.35}$/38.8$_{\pm0.69}$ \\
\rowcolor[gray]{0.8}w/ \algo(ours)  & \textbf{51.6}$_{\pm0.30}$/\textbf{41.8}$_{\pm0.61}$  & \textbf{60.3}$_{\pm0.75}$/\textbf{53.5}$_{\pm0.91}$ & \textbf{55.5}$_{\pm0.63}$/\textbf{46.5}$_{\pm0.62}$ \\
\botrule
\end{tabular}
\end{center}
\end{table}

To evaluate whether our method \algo\ performs balanced prediction for all classes, we measure its performance using Geometric Mean scores (GM) of class-wise accuracy. The results in~\Cref{tab5} demonstrate that the proposed algorithm yields the best and most balanced performance in all classes. Additionally, \algo\ achieves more significant performance improvement on the large dataset (CIFAR-100-LT).

\begin{table}[!h]
\begin{center}
\caption{Results of GM(\%) under the main setting.}\label{tab5}%
\begin{tabular}{lccc}
\toprule
  & CIFAR-10-LT  & SVHN-LT &  CIFAR-100-LT\\
\midrule
Algorithm & $\gamma = 100, \beta = 20\%$ & $\gamma = 100, \beta = 20\%$ & $\gamma = 20, \beta = 40\%$ \\
\midrule
FixMatch & 62.0 & 87.3 & 38.5  \\
w/ CReST+PDA & 74.4 & 88.6 & 42.3  \\
w/ DARP & 71.5 & 87.6 & 40.4  \\
w/ DARP+cRT  & 76.7 & 89.8 & 47.0  \\
w/ ABC  & 80.5 & 91.8 & 49.0 \\
\rowcolor[gray]{0.8} w/ \algo(ours)  & \textbf{81.9} & \textbf{93.4} & \textbf{54.0}   \\
\botrule
\end{tabular}
\end{center}
\end{table}

\subsection{How Does Pseudo-Label Distribution Impact the Performance?}

\begin{figure}[!h]%
\begin{subfigure}[b]{0.46\textwidth}
    \includegraphics[width=\textwidth]{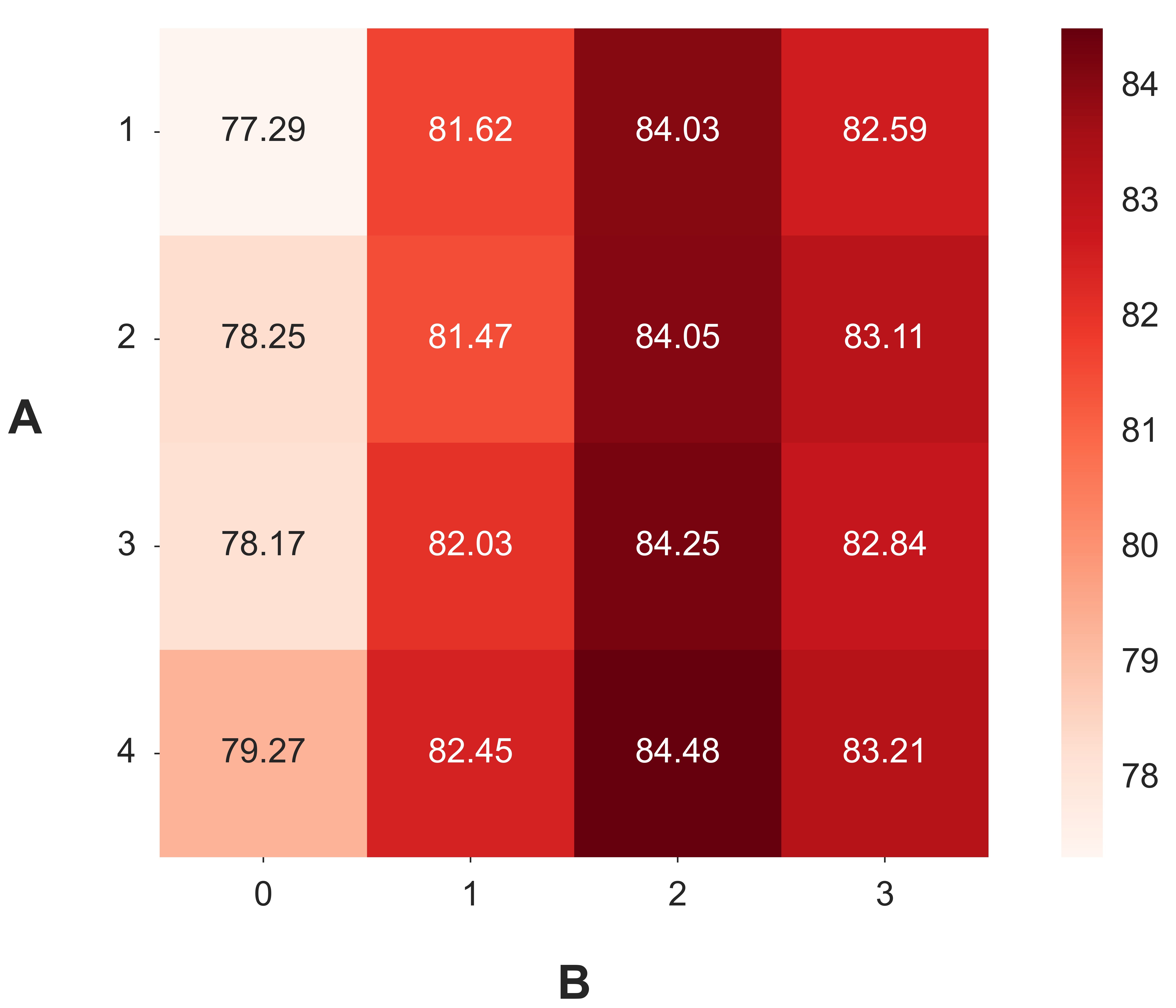}
    \caption{Overall accuracy}\label{Overall accuracy}
\end{subfigure}
\hfill
\begin{subfigure}[b]{0.46\textwidth}
    \includegraphics[width=\textwidth]{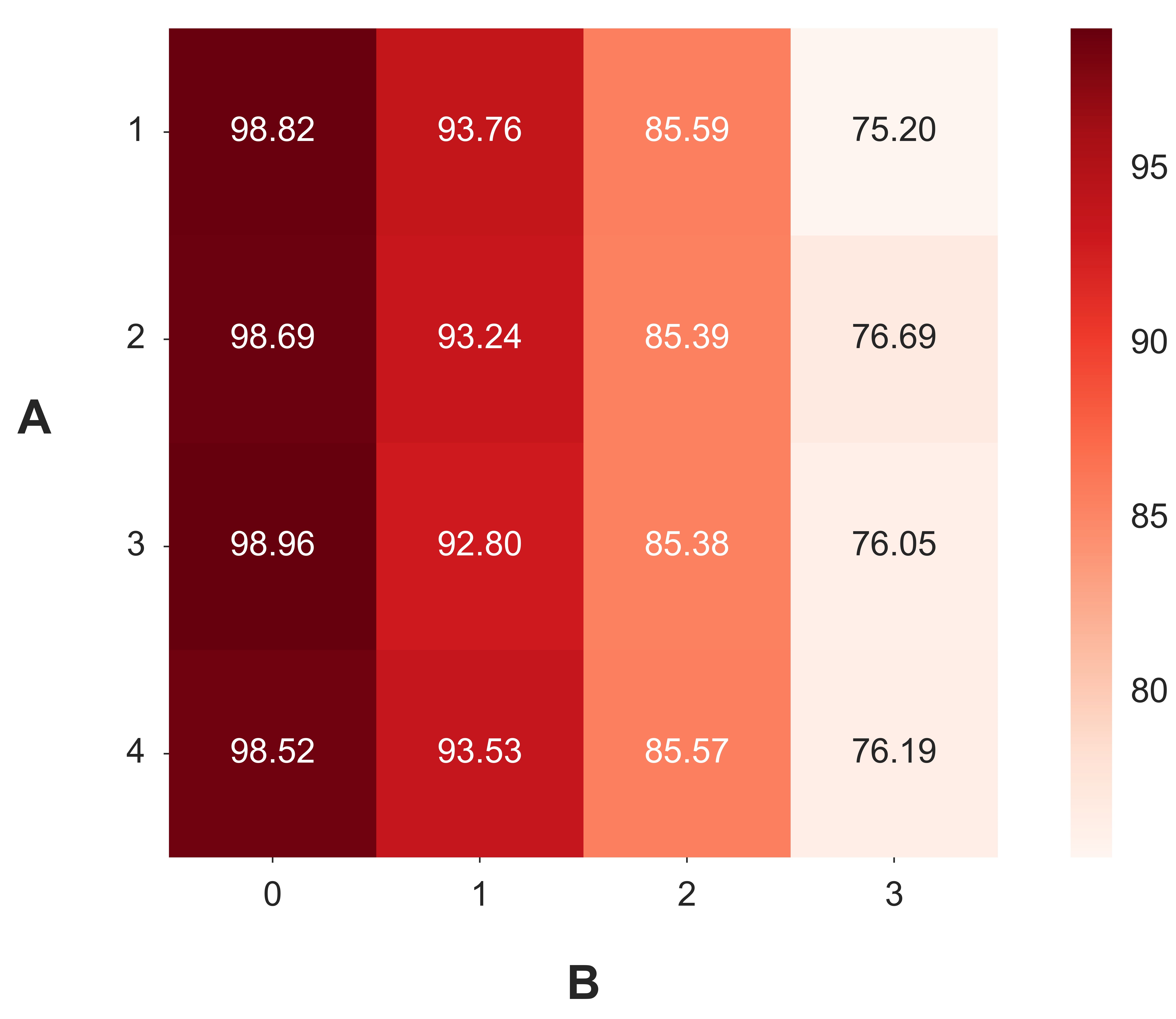}
    \caption{Top-5 accuracy of pseudo-labels}\label{Top-5}
\end{subfigure}
\caption{The impact of values of $A$ and $B$ on CIFAR-10-LT under class imbalance ratio 100 with 20\% of labels available. (a) Overall accuracy(\%) of \algo; (b) Top-5 accuracy(\%) of pseudo-labels of the teacher after transformation.}\label{fig:AB}
\end{figure}

Recall that we use hyper-parameters $A$ and $B$ to control the distribution of pseudo-labels by $\tau(\hat{y}) =A \cdot \boldsymbol{\alpha}_{\hat{y}} + B$, we now analyze their influence on the performance. The results are reported in \Cref{fig:AB}. We find that $B$ impacts the performance much larger than $A$, which coincides with our intuition because $\boldsymbol{\alpha}_{\hat{y}} < 1$. It achieves comparable results by setting $A \in \{1,2,3\}$. But unlike $A$, $B = 2$ yields better performance than other values in our experiments. When setting $A > 3$ and $B > 3$, test accuracy severely deteriorates because of the heavy bias towards minority classes in the pseudo-label distribution.

Interestingly, we find the balancedness of pseudo-label distribution matters much more than the accuracy of pseudo-labels in transfer. As $B$ increases, the top-5 accuracy of pseudo-labels is impaired while the overall accuracy remains competitive. This indicates that class imbalance hurts the performance more than inaccurate pseudo-labels in our approach.

\begin{figure}[!h]%
    \includegraphics[width=1\textwidth]{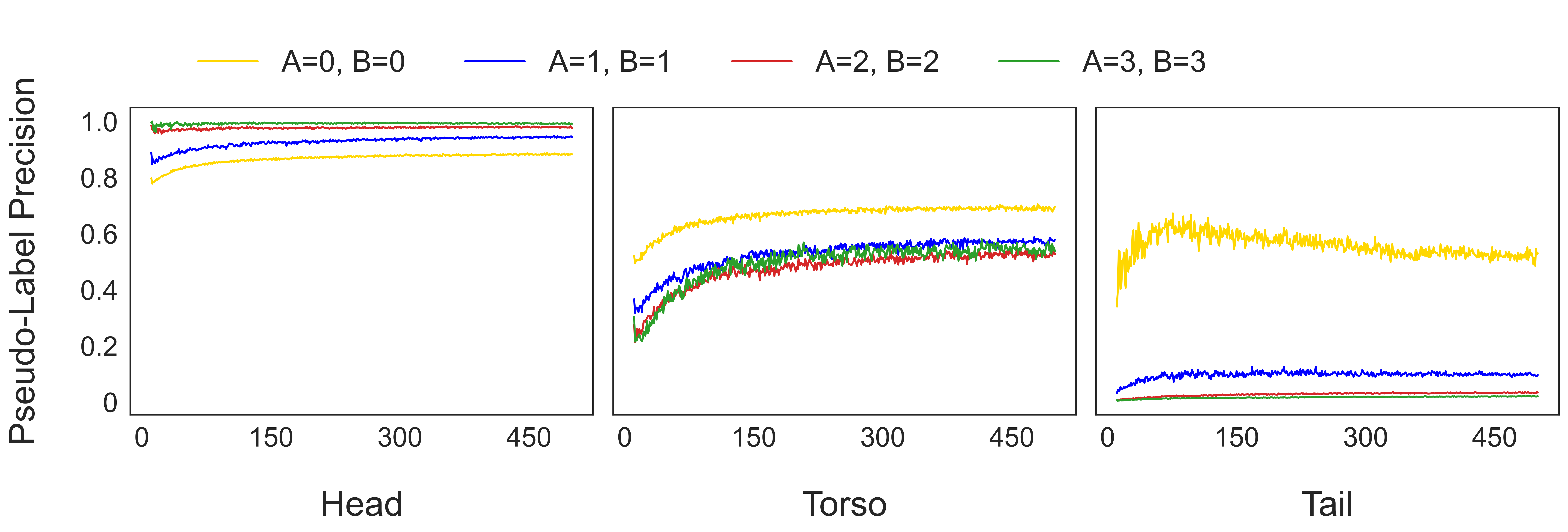}
    \caption{
		Comparison of pseudo-label precision by varying the values of A and B on CIFAR-10-LT under class imbalance ratio 100 with 20\% of labels available. The x-axis is the number of epochs, and the y-axis is the precision. Classes are divided into head (\{0, 1, 2\}), torso (\{3, 4, 5, 6\}) and tail (\{7, 8, 9\}). 
    } \label{P}
\end{figure}

\begin{figure}[!h]%
    \includegraphics[width=1\textwidth]{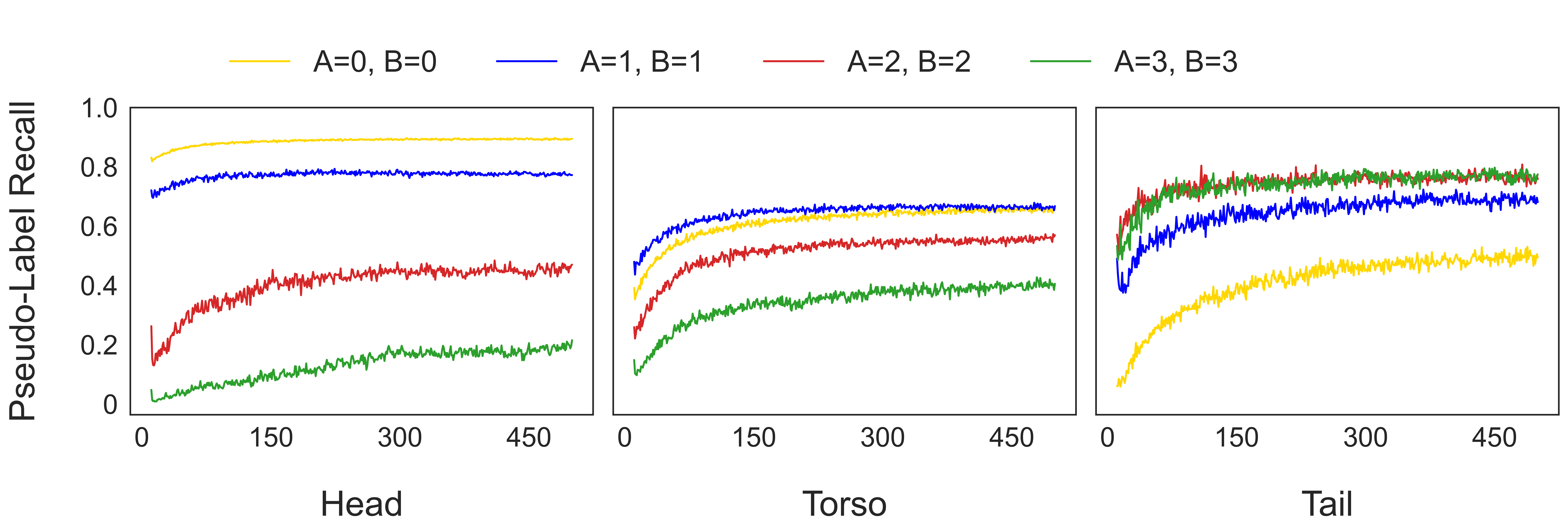}
    \caption{
		Comparison of pseudo-label recall by varying the value of A and B on CIFAR-10-LT under class imbalance ratio 100 with 20\% of labels. The x-axis is the number of epochs, and the y-axis is the recall. 
    } \label{R}
\end{figure}

\begin{figure}[!t]%
    \includegraphics[width=1\textwidth]{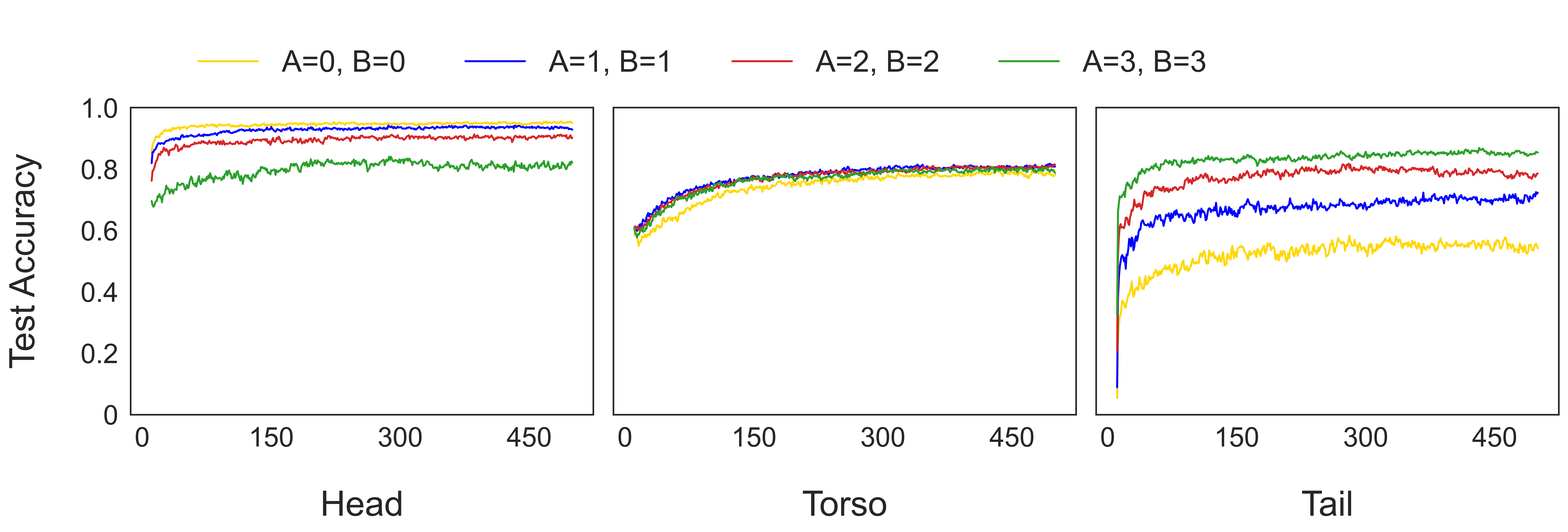}
    \caption{
		Comparison of test accuracy by varying the values of A and B on CIFAR-10-LT under class imbalance ratio 100 with 20\% of labels. The x-axis is the number of epochs, and the y-axis is the test accuracy.
    } \label{T}
\end{figure}

To better understand this phenomenon, we investigate the impact of logit scaling parameters $A$ and $B$ on the quality of pseudo-labels for head, torso, and tail classes separately. 
As illustrated in \Cref{P}, $A=0, B=0$ reveals superb performance with high precision. However, in \Cref{R}, it shows the worst recall in the tail class. Since $A=0, B=0$ means that pseudo-labels are from the conventional SSL model which is biased to the head class, transferring their distribution to a target model does not help the training of tail classes, as shown in \Cref{T}.

Instead, by setting $A=2, B=2$, it achieves the best performance in overall and tail-class accuracy as reported in \Cref{Overall accuracy} and \Cref{T}. Notably, it produces high recall yet low precision for tail classes in \Cref{R} and \Cref{P}. This observation confirms our suspicion that the balancedness of pseudo-label distribution requires more attention than the accuracy of pseudo-labels in knowledge transfer.

\subsection{Better Understanding of \algo}

\begin{figure}[h]%
\begin{subfigure}[b]{0.42\textwidth}
    \includegraphics[width=\textwidth]{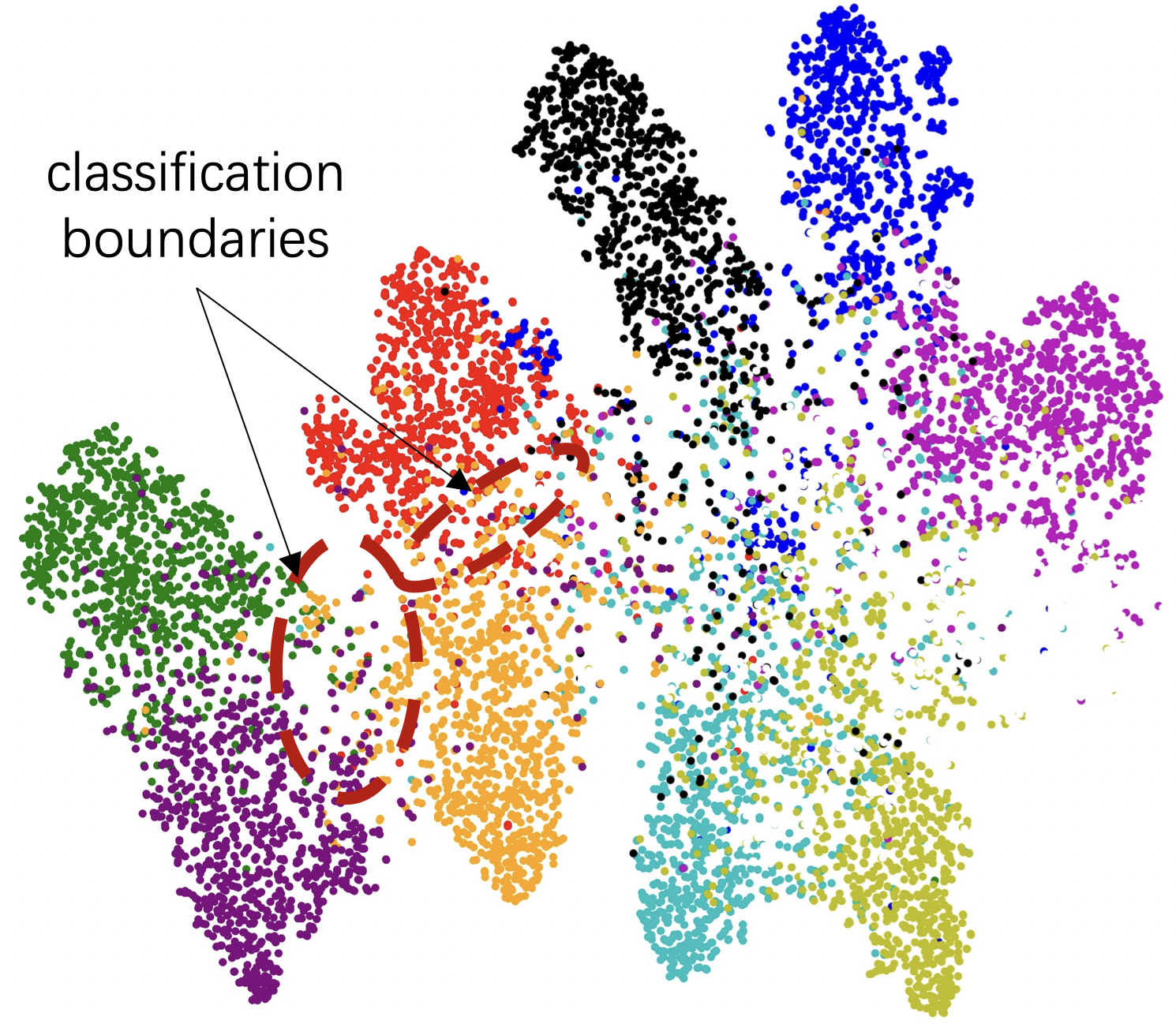}
    \caption{ABC}
\end{subfigure}
\hfill
\begin{subfigure}[b]{0.42\textwidth}
    \includegraphics[width=\textwidth]{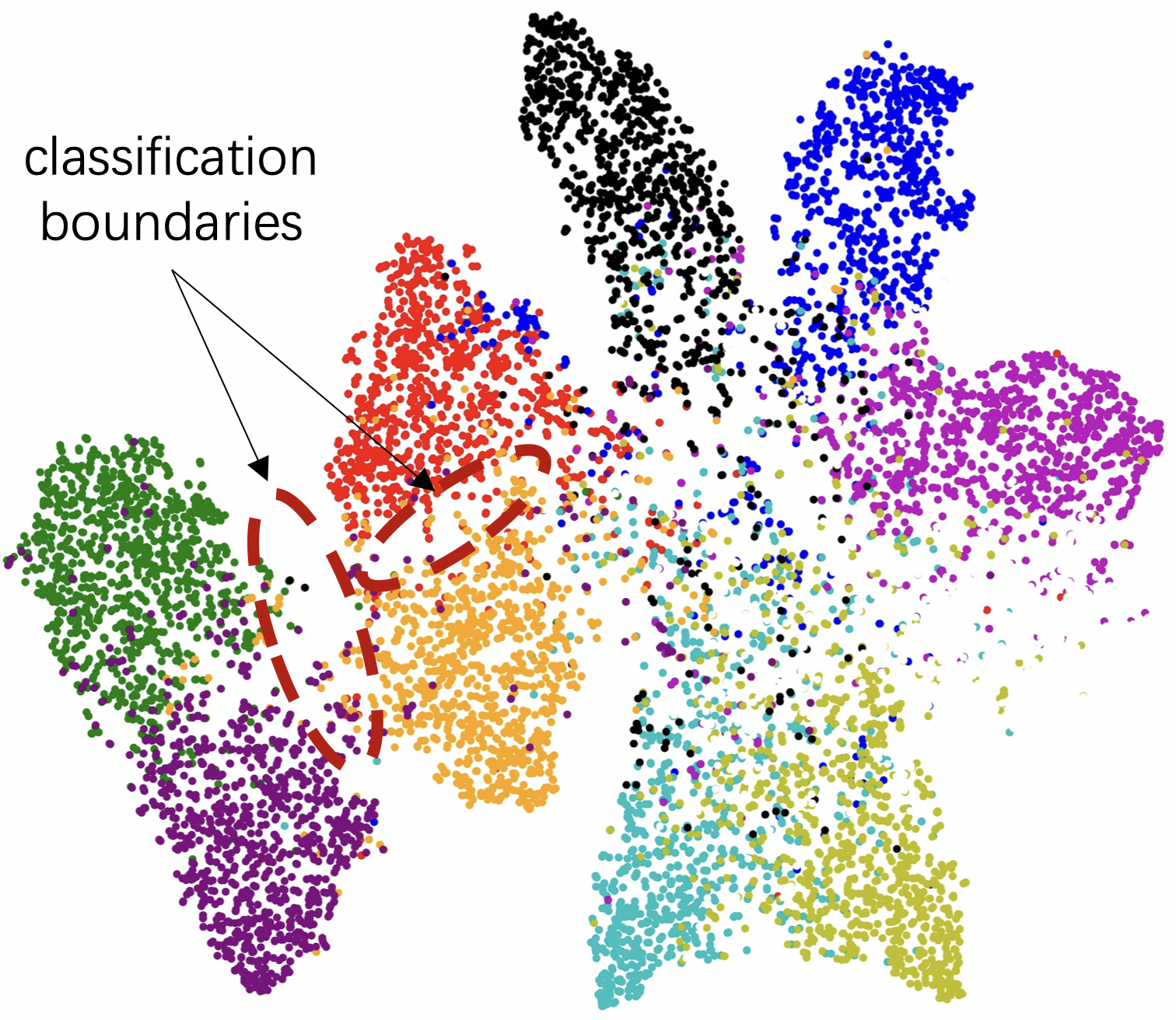}
    \caption{\algo}
\end{subfigure}
\caption{Results of t-SNE for ABC \citep{lee2021abc} and our \algo.}\label{tsne}
\end{figure}

We analyze \algo\ from representation and classification perspectives on CIFAR-10-LT under the main setting. 
First, we compare the learned representations by ABC and our \algo\ via t-distributed stochastic neighbor embedding (t-SNE) \citep{van2008visualizing} in \Cref{tsne}. It can be seen that \algo\ has more clear classification boundaries than ABC, which demonstrates that \algo\ can distinguish the difference between classes with better representation learning.

\begin{figure}[h]%
\begin{subfigure}[b]{0.45\textwidth}
    \includegraphics[width=\textwidth]{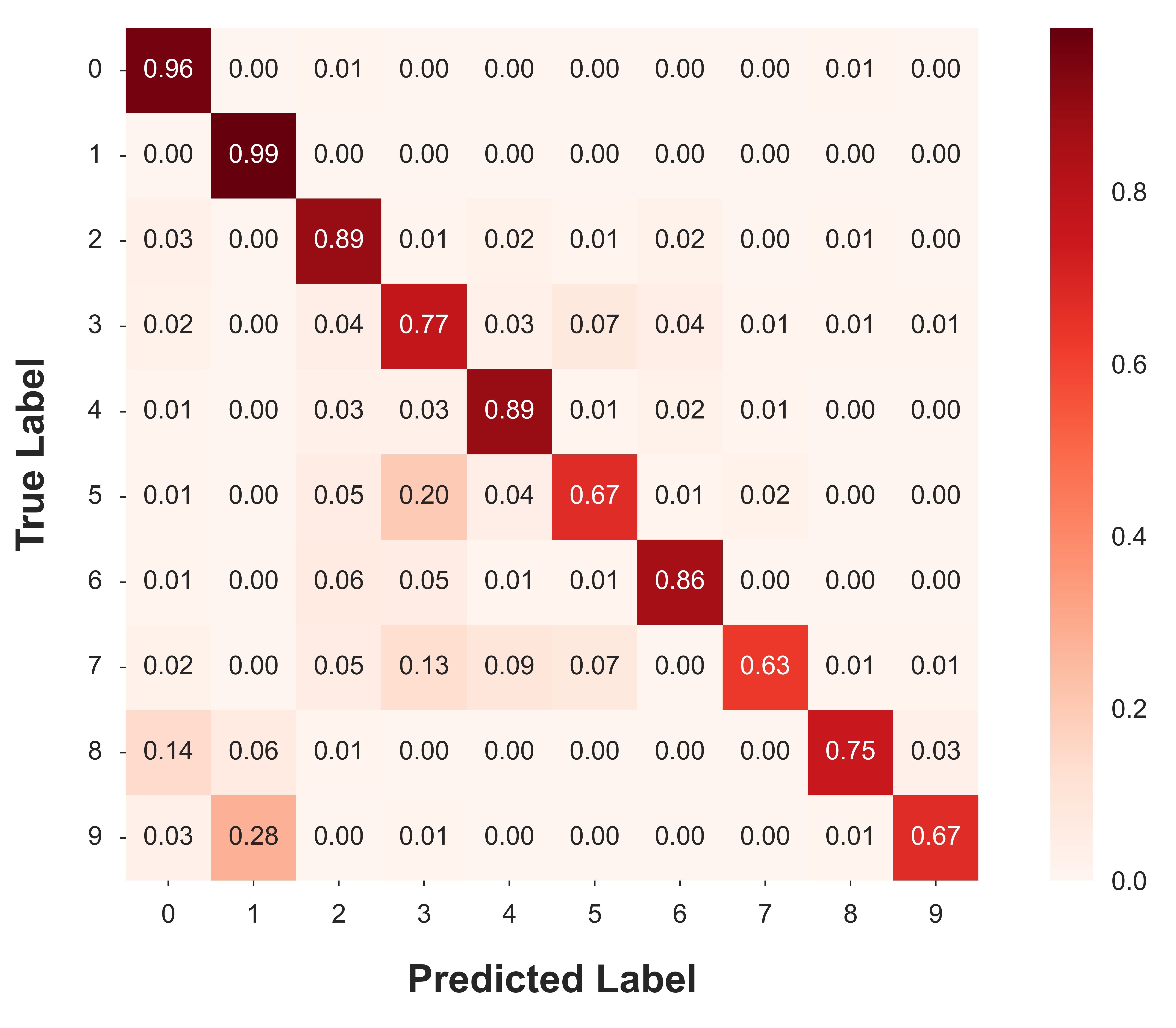}
    \caption{ABC}
\end{subfigure}
\hfill
\begin{subfigure}[b]{0.45\textwidth}
    \includegraphics[width=\textwidth]{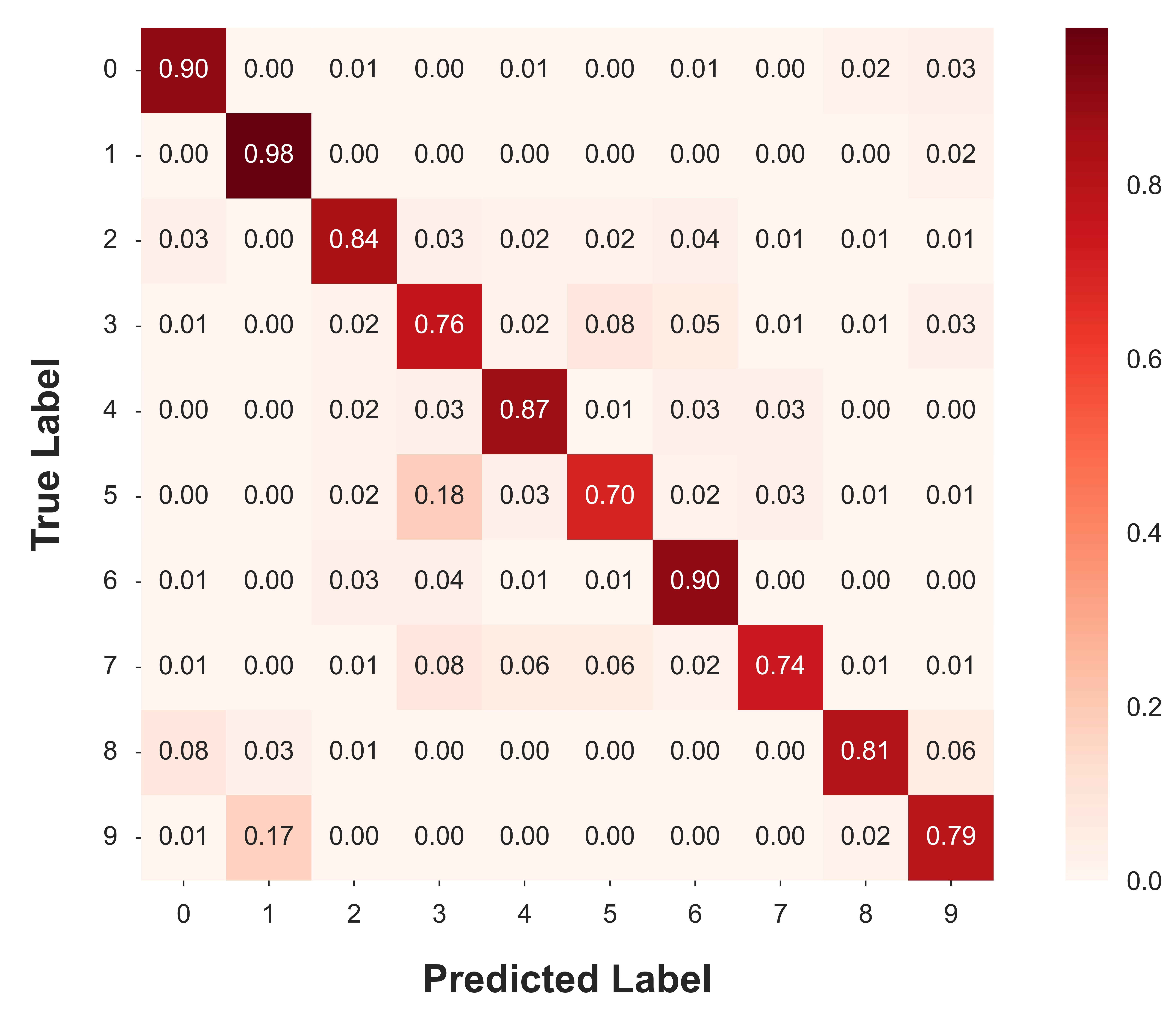}
    \caption{\algo}
\end{subfigure}
\caption{Confusion matrices of the prediction on the test set of CIFAR-10-LT.}\label{cm}
\end{figure}

Further, to analyze the classification results, we compare the confusion matrices of the prediction on the test set in \Cref{cm}. Each row represents the ground-truth label and each column represents the prediction by ABC or our \algo. The value in the $i$-th row and $j$-th column is the percentage of samples from the $i$-th class and predicted as the $j$-th class. From the results, we can see that our \algo\ performs better than ABC in the minority class. Moreover, it is observed that \algo\ might misclassify some majority-class samples as the minority-class ones. 

\textbf{Ablation studies.} We conduct ablation studies on important parts of our approach under the main setting. 

First, this paper applies logit transformation with $A=0, B=1$ to the teacher model's prediction on unlabeled data for better performance of the teacher model. By removing logit transformation, the overall accuracy and minority-class accuracy under the main setting turn out to be 83.66\% (-0.64\%) and 77.54\% (-4.66\%) on CIFAR-10-LT, respectively.

Second, we modify the distribution-aware cross-entropy for the labeled data to the common cross-entropy loss, leading to 83.41\% (-0.89\%) and 80.28\% (-1.92\%) of the overall accuracy and minority-class accuracy. The marginal decline of the performance verifies the effectiveness of the learning through imitation approach.

Finally, we remove the sample mask on unlabeled data of the student model, which means all unlabeled data is used to imitate the teacher. The experiment shows that removing the sample mask decreases the performance slightly, i.e., 83.33\% (-0.97\%) and 79.60\% (-2.59\%) for overall and minority-class accuracy respectively. This demonstrates the advantage of selecting more accurate pseudo-labels for the student model.

\subsection{Comparison with Two-stage Training}

We further compare \algo\ with two-stage training under the main setting in Table~\ref{tab6}. In the two-stage training, we first train a FixMatch model as the teacher and then guide a student model by the teacher with our \algo.
We find that not only can our \algo\ save training cost, but it is also better trained than the two-stage approach. This agrees with our expectation that, in two-stage training, the student lays more emphasis on the minority class, while fitting the pseudo-label distribution does not necessarily improve feature learning. Fortunately, two branches in \algo\ share the feature learning backbone, which can improve the backbone and classifiers simultaneously.

To further show that double branches of \algo\ can both improve feature learning, we stop propagating the gradients of the student branch from affecting the feature learning backbone. In this way, only the teacher model trains the feature extractor network, which is similar to the two-stage training. From the result, we see \algo\ performs better, showing that the student can further enhance the feature learning by sharing the backbone with the teacher model.

\begin{table}[h]
\begin{center}
\caption{Performance comparison of overall accuracy(\%)/minority-class accuracy(\%) with two-stage training and \algo \textminus.}\label{tab6}%
\begin{tabular}{lccc}
\toprule
  & CIFAR-10-LT  & SVHN-LT &  CIFAR-100-LT\\
\midrule
Method & $\gamma = 100, \beta = 20\%$ & $\gamma = 100, \beta = 20\%$ & $\gamma = 20, \beta = 40\%$ \\
\midrule
Two-stage & 80.6/75.0 & 91.6/89.6 & 57.1/47.6 \\
 \algo \textminus \footnotemark[1] & 83.2/80.8 & 92.6/92.1 & 57.8/50.3 \\
\rowcolor[gray]{0.8} \algo\  & \textbf{84.3}/\textbf{82.2} & \textbf{93.4}/\textbf{92.5} & \textbf{58.5}/\textbf{50.3} \\
\botrule
\end{tabular}
\footnotetext[1]{The student does not propagate its gradients to update the feature extractor.}
\end{center}
\end{table}

\section{Conclusion}
We introduce \algo, a new method for LTSSL. \algo\ (1) learns from a more class-balanced label distribution to improve the minority-class generalization and  (2) partitions the parameter space, enabling transfer via weight sharing of the transformed knowledge learned by the conventional SSL model. Extensive experiments on CIFAR-10-LT, SVHN-LT, and CIFAR-100-LT datasets show that \algo\ outperforms state-of-the-art methods on the minority class by a large margin. In the sequel, it would be interesting to extend \algo\ to more established SSL methods.


\bibliography{sn-bibliography}

\begin{thebibliography}{33}
\providecommand{\natexlab}[1]{#1}
\providecommand{\url}[1]{{#1}}
\providecommand{\urlprefix}{URL }
\providecommand{\doi}[1]{\url{https://doi.org/#1}}
\providecommand{\eprint}[2][]{\url{#2}}
 \bibcommenthead

\bibitem[{Amodei et~al(2016)Amodei, Ananthanarayanan, Anubhai, Bai, Battenberg,
  Case, Casper, Catanzaro, Cheng, Chen et~al}]{amodei2016deep}
Amodei D, Ananthanarayanan S, Anubhai R, et~al (2016) Deep speech 2: End-to-end
  speech recognition in english and mandarin. In: International Conference on
  Machine Learning, pp 173--182

\bibitem[{Berthelot et~al(2019{\natexlab{a}})Berthelot, Carlini, Cubuk,
  Kurakin, Sohn, Zhang, and Raffel}]{berthelot2019remixmatch}
Berthelot D, Carlini N, Cubuk ED, et~al (2019{\natexlab{a}}) Remixmatch:
  Semi-supervised learning with distribution matching and augmentation
  anchoring. In: International Conference on Learning Representations

\bibitem[{Berthelot et~al(2019{\natexlab{b}})Berthelot, Carlini, Goodfellow,
  Papernot, Oliver, and Raffel}]{berthelot2019mixmatch}
Berthelot D, Carlini N, Goodfellow I, et~al (2019{\natexlab{b}}) Mixmatch: A
  holistic approach to semi-supervised learning. Advances in Neural Information
  Processing Systems 32:5050--5060

\bibitem[{Branco et~al(2016)Branco, Torgo, and Ribeiro}]{branco2016survey}
Branco P, Torgo L, Ribeiro RP (2016) A survey of predictive modeling on
  imbalanced domains. ACM Computing Surveys 49(2):1--50

\bibitem[{van Engelen and Hoos(2020)}]{DBLP:journals/ml/EngelenH20}
van Engelen JE, Hoos HH (2020) A survey on semi-supervised learning. Machine
  Learning 109(2):373--440

\bibitem[{Gao et~al(2017)Gao, Xing, Xie, Wu, and Geng}]{gao2017deep}
Gao BB, Xing C, Xie CW, et~al (2017) Deep label distribution learning with
  label ambiguity. IEEE Transactions on Image Processing 26(6):2825--2838

\bibitem[{Geng(2016)}]{geng2016label}
Geng X (2016) Label distribution learning. IEEE Transactions on Knowledge and
  Data Engineering 28(7):1734--1748

\bibitem[{Guo et~al(2020)Guo, Zhang, Jiang, Li, and Zhou}]{guo2020safe}
Guo LZ, Zhang ZY, Jiang Y, et~al (2020) Safe deep semi-supervised learning for
  unseen-class unlabeled data. In: International Conference on Machine
  Learning, pp 3897--3906

\bibitem[{He and Garcia(2009)}]{he2009learning}
He H, Garcia EA (2009) Learning from imbalanced data. IEEE Transactions on
  Knowledge and Data Engineering 21(9):1263--1284

\bibitem[{He et~al(2016)He, Zhang, Ren, and Sun}]{he2016deep}
He K, Zhang X, Ren S, et~al (2016) Deep residual learning for image
  recognition. In: Proceedings of the IEEE/CVF Conference on Computer Vision
  and Pattern Recognition, pp 770--778

\bibitem[{He et~al(2021)He, Wu, and Wei}]{he2021distilling}
He YY, Wu J, Wei XS (2021) Distilling virtual examples for long-tailed
  recognition. In: Proceedings of the IEEE/CVF International Conference on
  Computer Vision, pp 235--244

\bibitem[{Iscen et~al(2021)Iscen, Araujo, Gong, and Schmid}]{iscen2021cbd}
Iscen A, Araujo A, Gong B, et~al (2021) Class-balanced distillation for
  long-tailed visual recognition. In: The British Machine Vision Conference

\bibitem[{Kang et~al(2020)Kang, Xie, Rohrbach, Yan, Gordo, Feng, and
  Kalantidis}]{kang2019decoupling}
Kang B, Xie S, Rohrbach M, et~al (2020) Decoupling representation and
  classifier for long-tailed recognition. In: International Conference on
  Learning Representations

\bibitem[{Kim et~al(2020)Kim, Hur, Park, Yang, Hwang, and
  Shin}]{kim2020distribution}
Kim J, Hur Y, Park S, et~al (2020) Distribution aligning refinery of
  pseudo-label for imbalanced semi-supervised learning. Advances in Neural
  Information Processing Systems 33:14,567--14,579

\bibitem[{Kingma and Ba(2015)}]{kingma2015adam}
Kingma DP, Ba J (2015) Adam: A method for stochastic optimization. In:
  International Conference on Learning Representations

\bibitem[{Lee et~al(2021)Lee, Shin, and Kim}]{lee2021abc}
Lee H, Shin S, Kim H (2021) Abc: Auxiliary balanced classifier for
  class-imbalanced semi-supervised learning. Advances in Neural Information
  Processing Systems 34:7082--7094

\bibitem[{Li et~al(2019)Li, Wang, Wei, and Tu}]{DBLP:conf/aaai/LiWWT19}
Li Y, Wang H, Wei T, et~al (2019) Towards automated semi-supervised learning.
  In: Proceedings of the AAAI Conference on Artificial Intelligence, pp
  4237--4244

\bibitem[{Liu et~al(2019)Liu, Miao, Zhan, Wang, Gong, and Yu}]{liu2019large}
Liu Z, Miao Z, Zhan X, et~al (2019) Large-scale long-tailed recognition in an
  open world. In: Proceedings of the IEEE/CVF Conference on Computer Vision and
  Pattern Recognition, pp 2537--2546

\bibitem[{Van~der Maaten and Hinton(2008)}]{van2008visualizing}
Van~der Maaten L, Hinton G (2008) Visualizing data using t-sne. Journal of
  Machine Learning Research 9(11)

\bibitem[{Menon et~al(2020)Menon, Jayasumana, Rawat, Jain, Veit, and
  Kumar}]{menon2020long}
Menon AK, Jayasumana S, Rawat AS, et~al (2020) Long-tail learning via logit
  adjustment. In: International Conference on Learning Representations

\bibitem[{Miyato et~al(2018)Miyato, Maeda, Koyama, and
  Ishii}]{miyato2018virtual}
Miyato T, Maeda Si, Koyama M, et~al (2018) Virtual adversarial training: a
  regularization method for supervised and semi-supervised learning. IEEE
  Transactions on Pattern Analysis and Machine Intelligence 41(8):1979--1993

\bibitem[{Ren et~al(2020)Ren, Yu, Ma, Zhao, Yi et~al}]{ren2020balanced}
Ren J, Yu C, Ma X, et~al (2020) Balanced meta-softmax for long-tailed visual
  recognition. Advances in Neural Information Processing Systems 33:4175--4186

\bibitem[{Sohn et~al(2020)Sohn, Berthelot, Carlini, Zhang, Zhang, Raffel,
  Cubuk, Kurakin, and Li}]{sohn2020fixmatch}
Sohn K, Berthelot D, Carlini N, et~al (2020) Fixmatch: Simplifying
  semi-supervised learning with consistency and confidence. Advances in Neural
  Information Processing Systems 33:596--608

\bibitem[{Tarvainen and Valpola(2017)}]{tarvainen2017mean}
Tarvainen A, Valpola H (2017) Mean teachers are better role models:
  Weight-averaged consistency targets improve semi-supervised deep learning
  results. Advances in Neural Information Processing Systems 30:1195--1204

\bibitem[{Wang and Geng(2019)}]{wang2019classification}
Wang J, Geng X (2019) Classification with label distribution learning. In:
  International Joint Conference on Artificial Intelligence, pp 3712--3718

\bibitem[{Wei et~al(2021{\natexlab{a}})Wei, Sohn, Mellina, Yuille, and
  Yang}]{wei2021crest}
Wei C, Sohn K, Mellina C, et~al (2021{\natexlab{a}}) Crest: A class-rebalancing
  self-training framework for imbalanced semi-supervised learning. In:
  Proceedings of the IEEE/CVF Conference on Computer Vision and Pattern
  Recognition, pp 10,857--10,866

\bibitem[{Wei and Li(2019)}]{weit2020tnnls}
Wei T, Li YF (2019) Does tail label help for large-scale multi-label learning?
  IEEE Transactions on Neural Networks and Learning Systems 31(7):2315--2324

\bibitem[{Wei et~al(2021{\natexlab{b}})Wei, Shi, Tu, and Li}]{Wei_2021_RoLT}
Wei T, Shi J, Tu W, et~al (2021{\natexlab{b}}) Robust long-tailed learning
  under label noise. CoRR abs/2108.11569

\bibitem[{Wei et~al(2022)Wei, Shi, Li, and Zhang}]{DBLP:conf/pakdd/WeiSLZ22}
Wei T, Shi J, Li Y, et~al (2022) Prototypical classifier for robust
  class-imbalanced learning. In: Proceedings of the Pacific-Asia Conference on
  Knowledge Discovery and Data Mining, pp 44--57

\bibitem[{Xiang et~al(2020)Xiang, Ding, and Han}]{xiang2020learning}
Xiang L, Ding G, Han J (2020) Learning from multiple experts: Self-paced
  knowledge distillation for long-tailed classification. In: European
  Conference on Computer Vision, pp 247--263

\bibitem[{Zagoruyko and Komodakis(2016)}]{zagoruyko2016wide}
Zagoruyko S, Komodakis N (2016) Wide residual networks. In: The British Machine
  Vision Conference

\bibitem[{Zhou et~al(2021)Zhou, Guo, Cheng, Li, and Pu}]{zhou2021step}
Zhou Z, Guo LZ, Cheng Z, et~al (2021) Step: Out-of-distribution detection in
  the presence of limited in-distribution labeled data. Advances in Neural
  Information Processing Systems 34:29,168--29,180

\bibitem[{Zhu et~al(2022)Zhu, Niu, Hua, and Zhang}]{beierxERM}
Zhu B, Niu Y, Hua XS, et~al (2022) Cross-domain empirical risk minimization for
  unbiased long-tailed classification. In: Proceedings of the AAAI Conference
  on Artificial Intelligence

\end{thebibliography}


\end{document}